# Automatic Removal of Marginal Annotations in Printed Text Document

Abdessamad Elboushaki*, Rachida Hannane, P. Nagabhushan, Mohammed Javed

Department of Studies in Computer Science, University of Mysore, Mysore,570006,India

**Abstract**

Recovering the original printed texts from a document with added handwritten annotations in the marginal area is one of the challenging problems, especially when the original document is not available**.** Therefore, this paper aims at salvaging automatically the original document from the annotated document by detecting and removing any handwritten annotations that appear in the marginal area of the document without any loss of information. Here a two stage algorithm is proposed, where in the first stage due to approximate marginal boundary detection with horizontal and vertical projection profiles, all of the marginal annotations along with some part of the original printed text that may appear very close to the marginal boundary are removed. Therefore as a second stage, using the connected components, a strategy is applied to bring back the printed text components cropped during the first stage. The proposed method is validated using a dataset of 50 documents having complex handwritten annotations, which gives an overall accuracy of 89.01% in removing the marginal annotations and 97.74% in case of retrieving the original printed text document.

## 1. Introduction

Adding annotation is the process or the act of writing critical commentary or explanatory notes into the printed text documents. Marking these annotations on the marginal area of printed text document is one of the common task that many people do whenever they read a document. These annotations may be some important observations or corrections marked in their own style which make the problem of recovering the original document very challenging, particularly when the original document is not available.

A reader can add different types of annotations in the marginal space which depends on the information being read by the reader. For example, the annotations can be a double line or a single vertical line marked on the boundary of the paragraph( in case of very interesting information), or it can be simply a question mark(when the reader does not understand the paragraph), or it can be in the form of more frequent annotations that are handwritten comments( in which the related idea is extrapolated  or the missed details are added). Fig 1.1(b) shows a sample document with different handwritten annotations added on the marginal area.

Therefore, automatic removal of handwritten annotations from the marginal area of the document becomes challenging especially when the annotations touch the marginal boundaries of the printed text. However, the problem of removing annotations arises when the copy of original document is not available and we have to depend on annotated copy of the original document. Thus, removing handwritten annotations from the marginal area of a printed text document still remains a major challenge in document analysis.

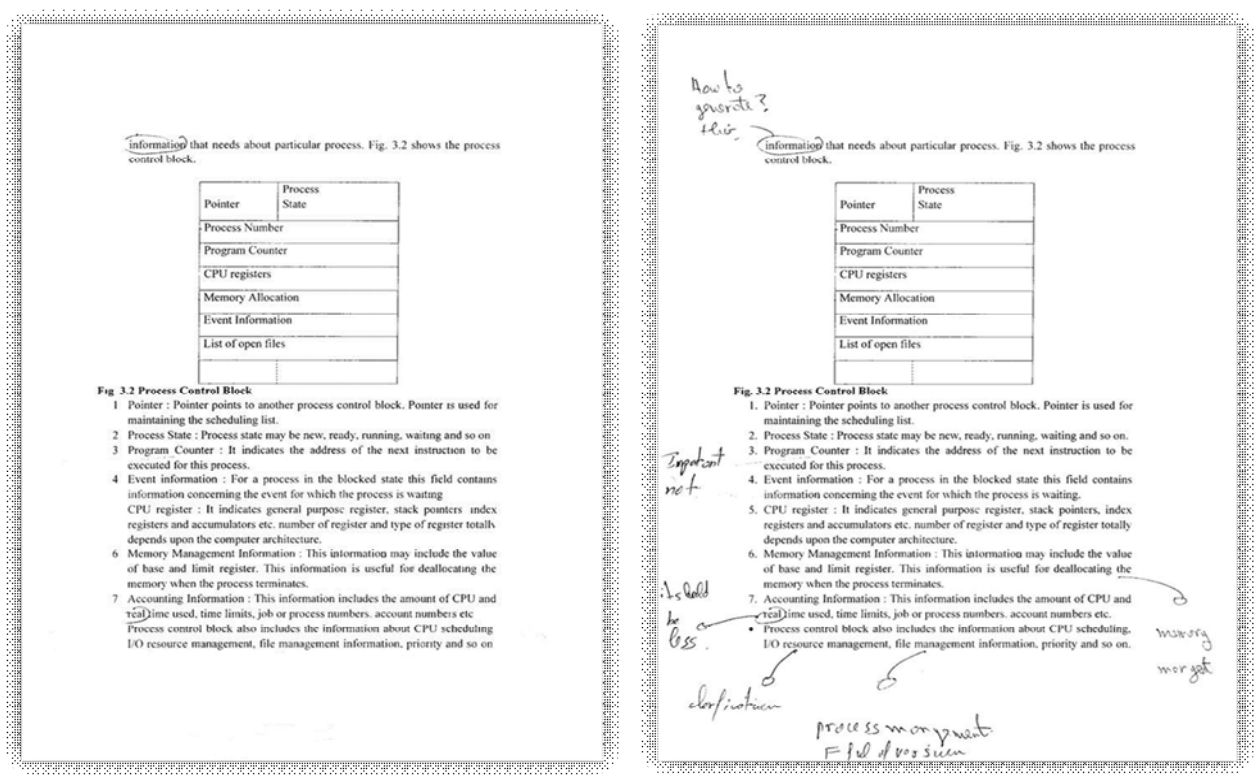

(a) Original Document (b) Annotated Document

* Corresponding author. Tel.:+919611025132.

*E-mail address*:abdessamad.elboushaki@gmail.com.

Fig 1.1 : Sample of digital document with different handwritten annotations

In this paper, we propose a method to automatically remove the marginal annotations in order to recover the original document in two stages, where in the first stage the approximate marginal boundary is detected and marginal annotations are removed using horizontal and vertical projection profiles. Because of the approximate marginal boundary detection, there is a possibility that some printed text are also removed which are recovered in the second stage using analysis of connected components. Rest of the paper is organized as follows: In Section II, we review some related works on printed text documents with handwritten annotations. In Section III, we detail the proposed model to salvage the original document. In Section IV, the experimental results of our proposed method are presented, and the last Section V concludes the paper.

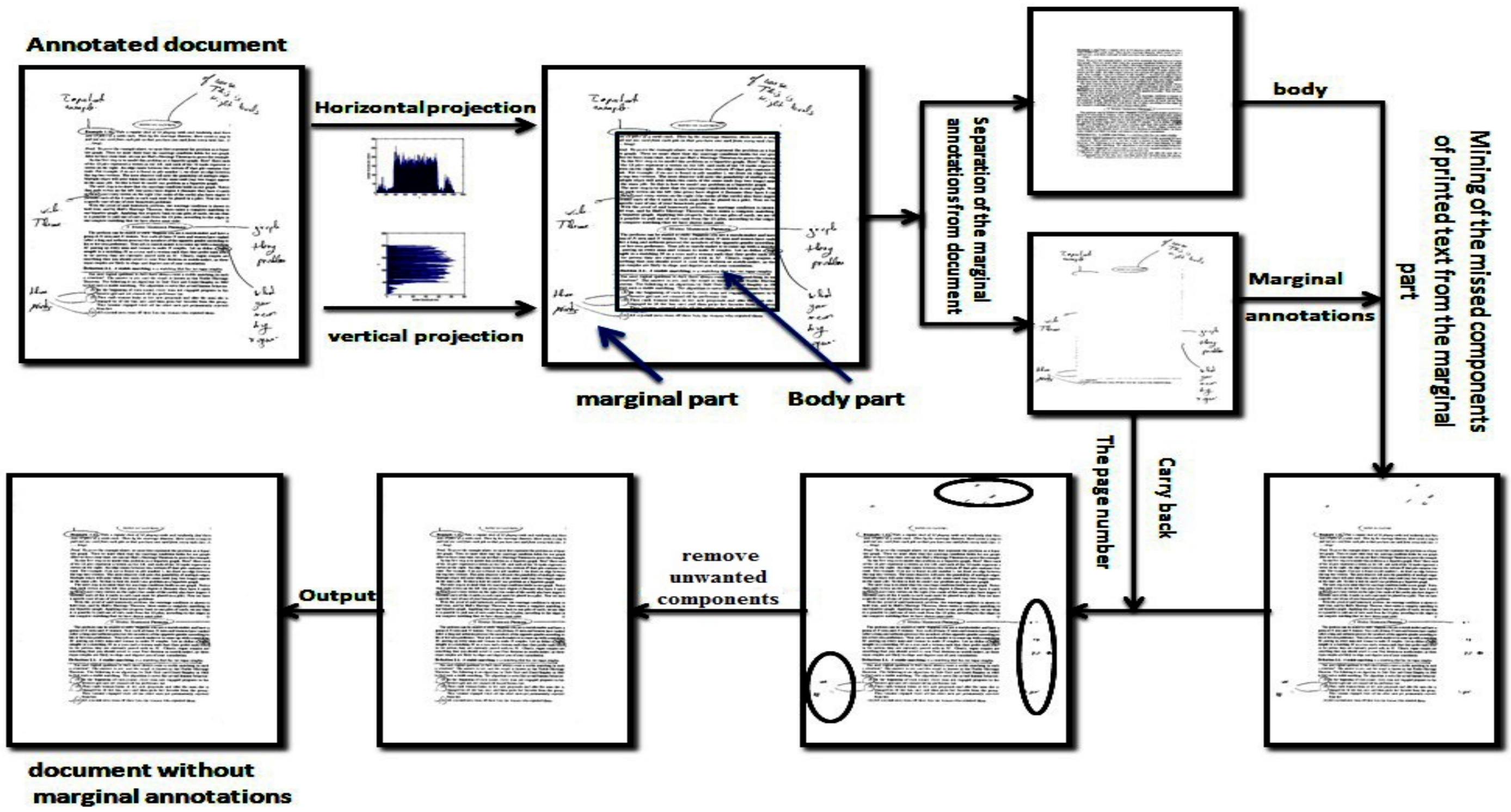

Fig 3.1 : Proposed Model

## 2. Related work

There are several methods proposed in the literature for extracting handwritten annotations from the document [1–6]. In [1], the authors use the original document images, in which the annotations are extracted by subtracting the original document from its annotated document. Their proposed method will work only under the assumption of availability of original document. Also, their subtraction method has the problem of image degradation between original document and the annotated document which may be caused due to use of different scanners.

There are some other methods [2–5] which do not require the original document to extract the annotations, in which annotations are extracted only from the annotated document images based on some predefined colours and shapes of annotations. Unfortunately these methods are limited to few shapes and some number of colours. However, in reality there is no limit on number of shapes and colours that may appear in the form of annotations.

In 2005, Ma and Guo [6] proposed a method for removing the annotations from marginal area of the document text. Their patented invention based on comparing the merged text lines to the regular pattern of the projection histograms, the printed text lines are discriminated from the handwritten annotations. This method can extract only the annotations which do not touch the printed text boundaries. However, in this paper, we present a method which can remove more complex annotations than the aforementioned methods and also without using the original document.

## 3. Proposed model

An overview of our proposed model is shown in Fig 3.1. The scanned document is pre-processed to remove noise coming out of document scanning. The approach used here is described in Peerawit and Kawtrakul [7]. Their method was applied from image processing technique called Sobel edge detection by using the edge density property of the noise and text areas. Later, a pre-processing to correct any skew present in the document images is applied. We use the method which is provided by Cao and Li in [8].

In this paper, we present a two stage algorithm for removing the handwritten marginal annotations. They are described as follows.

### *3.1 Marginal boundary detection and removal of annotations*

This stage determines the horizontal marginal (top and bottom) area and vertical marginal (Left and right side) area using respectively vertical and horizontal projection profiles.

In case of vertical marginal area, we plot the horizontal projection of the document and then a smoothing technique is applied to the histogram. In our experiment, a mask that gives a good result to determine vertical marginal is

$$smothing_mask = \frac{2x}{mean(a)} \tag{1}$$

where **x** is the length of the document and **a** is the density of the black pixels in the histogram.

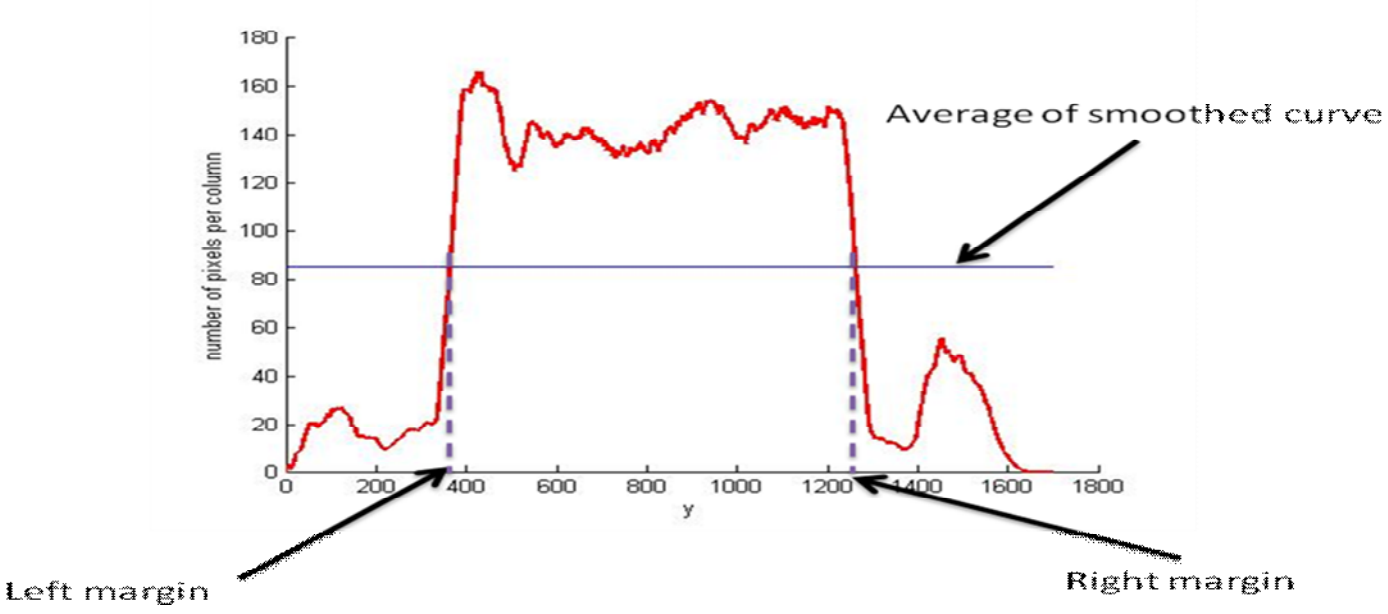

Fig 3.2 : Vertical margin detection

The two intersection points between the smoothing curve and the mean of the histogram are expected to be the vertical (left and right) margin. Fig 3.2 shows the detection of the vertical marginal area.

In detecting the horizontal marginal area, we use same approach as in the case of vertical margin detection. However, the smoothing mask used in the vertical projection is

$$smothing_mask = \frac{2y}{mean(b)} \tag{2}$$

where *y* is the width of the document and *b* is the density of the black pixels in the histogram.

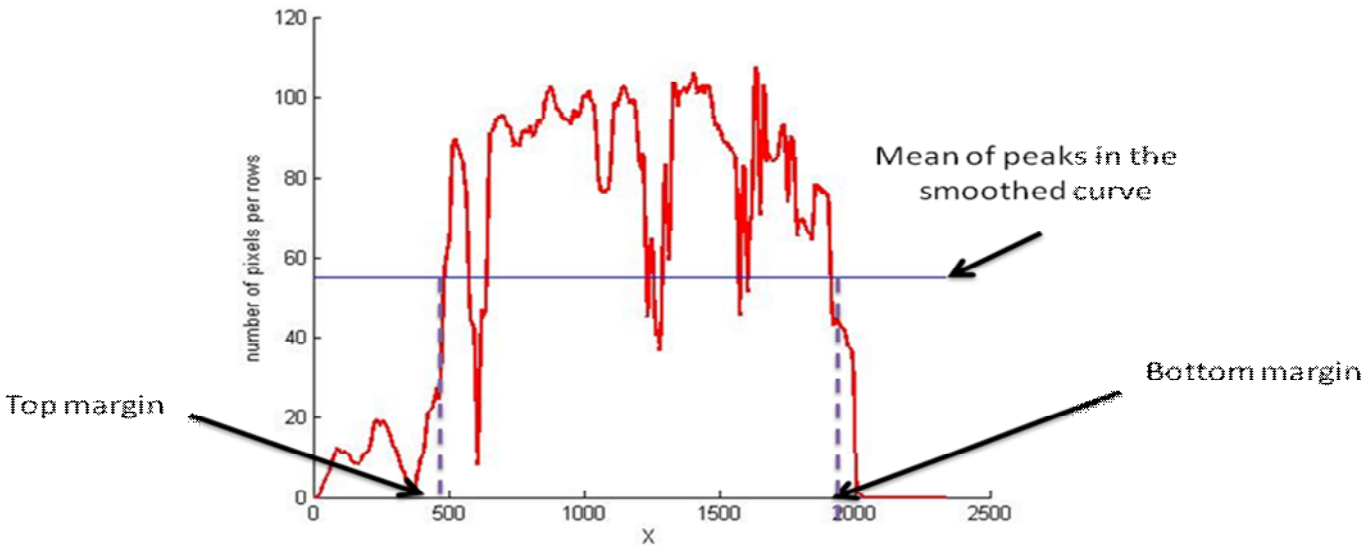

Fig 3.3 : Horizontal margin detection

From the left side to the right side of the histogram in Fig.3.3, the first and last intersections between the smoothing curve and the mean of the peaks in the smoothed histogram are identified as horizontal (top and bottom) margin. Fig 3.3 shows the detection of the horizontal marginal area.

After detecting the boundaries of the marginal side, the annotations from both horizontal and vertical marginal areas are removed in one stretch. Fig 3.4 demonstrates the stage of separation of marginal annotations from a sample annotated document. However, in this stage we observe that due to presence of marginal annotations, the marginal boundaries detected are approximate and this results in removal of some portion of printed text from the document. Therefore, these printed text have to be detected and brought back into the original document which is discussed in the second stage.

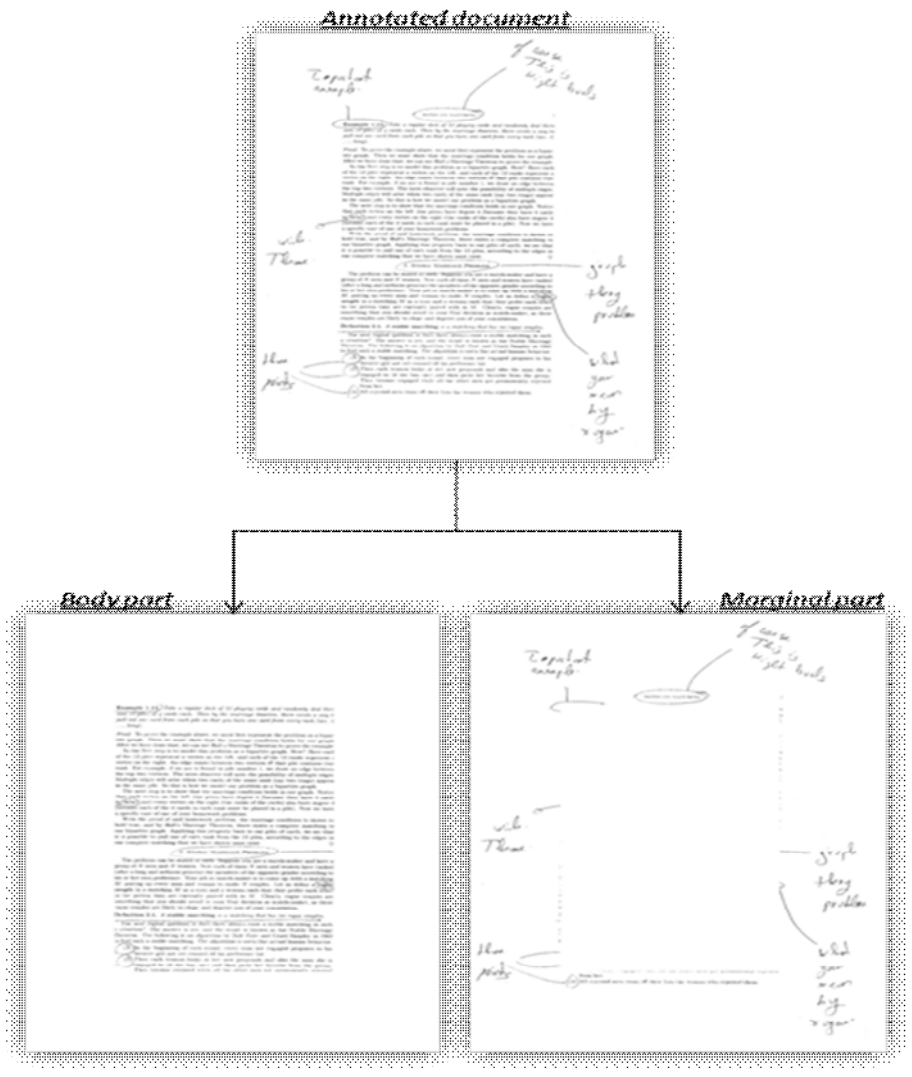

Fig 3.4 : Separation of body part and marginal area

### *3.2 Recovering printed text from the removed marginal annotations*

After the separation of marginal annotations from the annotated document, there is a possibility that some portions of printed text are also deleted because of approximate marginal boundary detection. Therefore we need to detect the printed text and bring it back to the original document. Fig 3.5 shows the possible cases that may occur in a document.

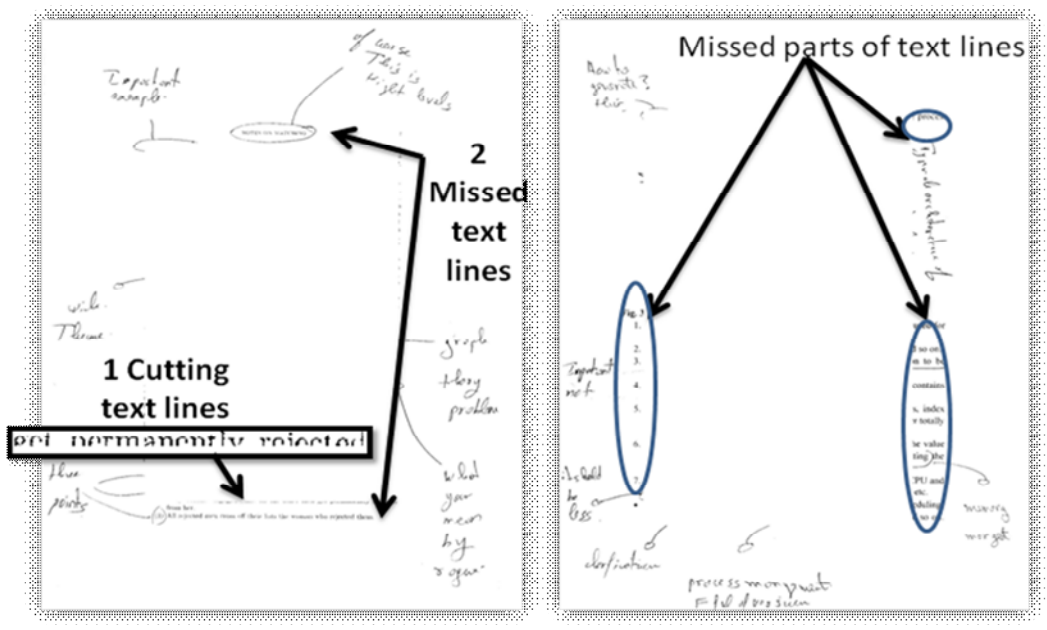

(a) (b)

Fig 3.5: Possible cases of printed text that should be brought back to the original document

#### *3.2.1 Broken text lines in the horizontal marginal area*

Due to approximate marginal boundary detection, some portion of the text lines get deleted from the main body and the remaining parts come along with marginal annotations.

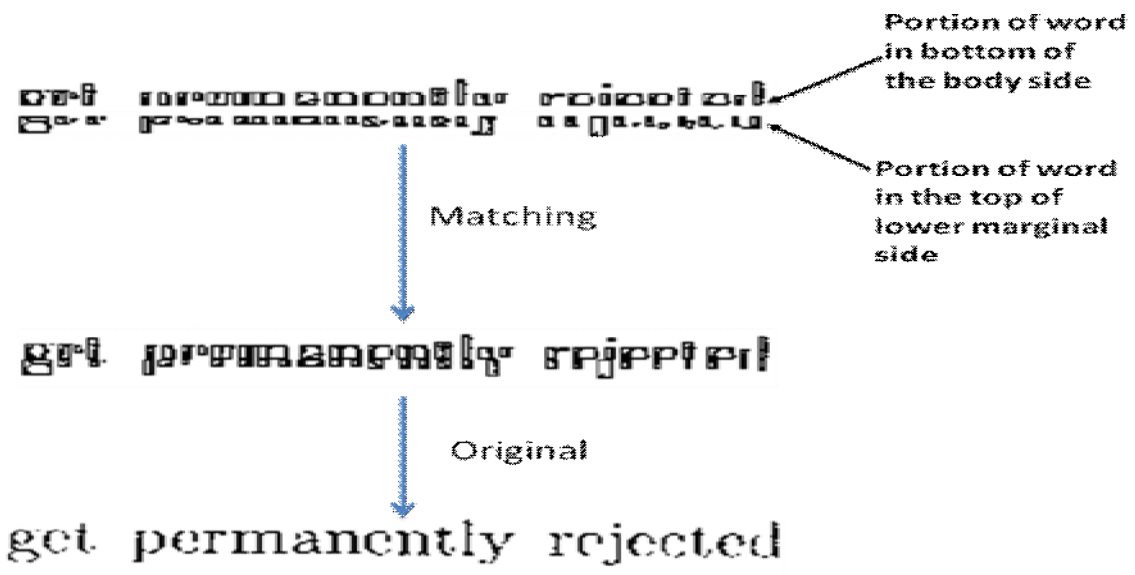

Fig 3.6: Detection of broken text lines

To recover the broken text lines, we use the concept of labelling algorithm to determine the connected components. We look for any connected component in the top of marginal area that touches with the corresponding connected component of top portion of the document body. If there exists such a case, provided that size of both the connected components do not exceed twice the character size of a printed text in a document, then that part of text line from the marginal side is brought back to the document body. Similar procedure is applied to bring back broken text lines from the bottom marginal area of the document. Fig 3.6 shows this procedure.

### *3.2.2 Missed text lines in the horizontal marginal area*

There may exist few words or small text lines that can be missed from the document body (for example chapter name at the top margin). In order to bring back these type of missed text lines, we apply the same concept of labelling algorithm in the marginal side and then we merge all the connected components that belong to the same text line subjected to the condition that they do not exceed twice of character size of the printed text.

Further we bring back those merged components from marginal side into body side. Fig 3.7 demonstrates retrieving of missed text lines.

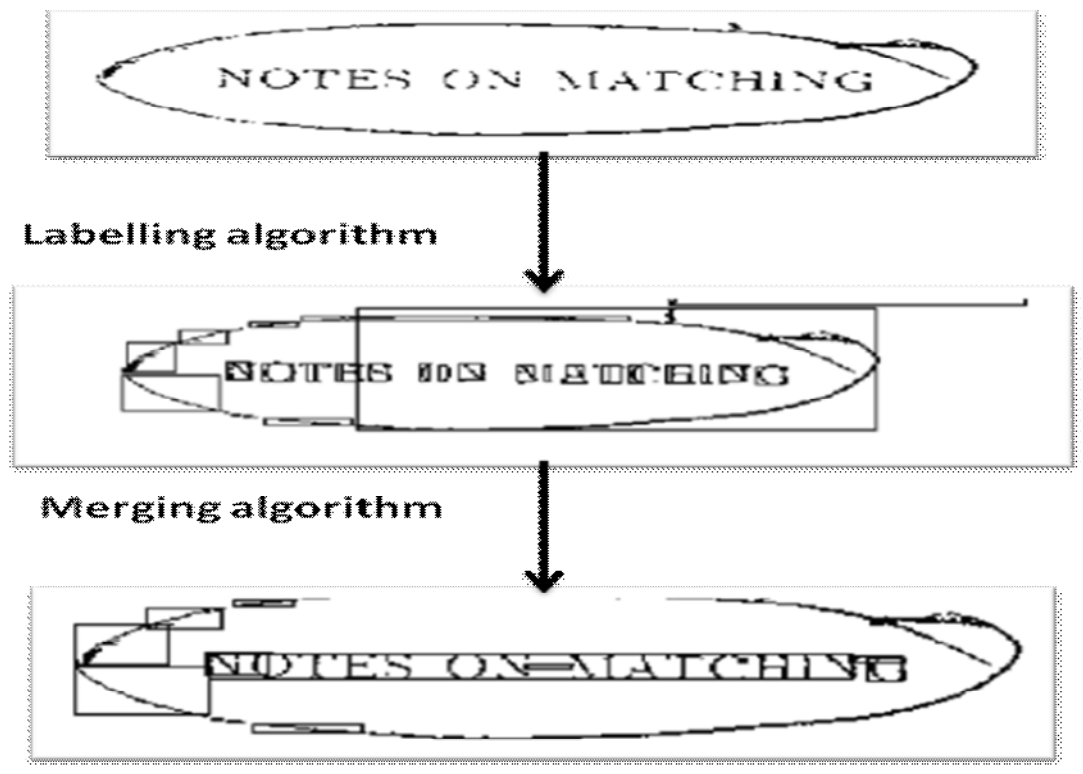

Fig 3.7 : Retrieving missed text lines

### *3.2.3 Missed part of the text lines in the vertical marginal area*

Fig 3.5 (b) shows some missed parts of the printed text lines in the vertical marginal area. To retrieve them, we check the boundaries of every text line in the body of the document image to find if there is any connected component that is in the same position as that of a text line, and it shares the same properties as of printed text (character size, character space and position ). Fig 3.8 shows the case of retrieving words which were deleted from the vertical marginal area.

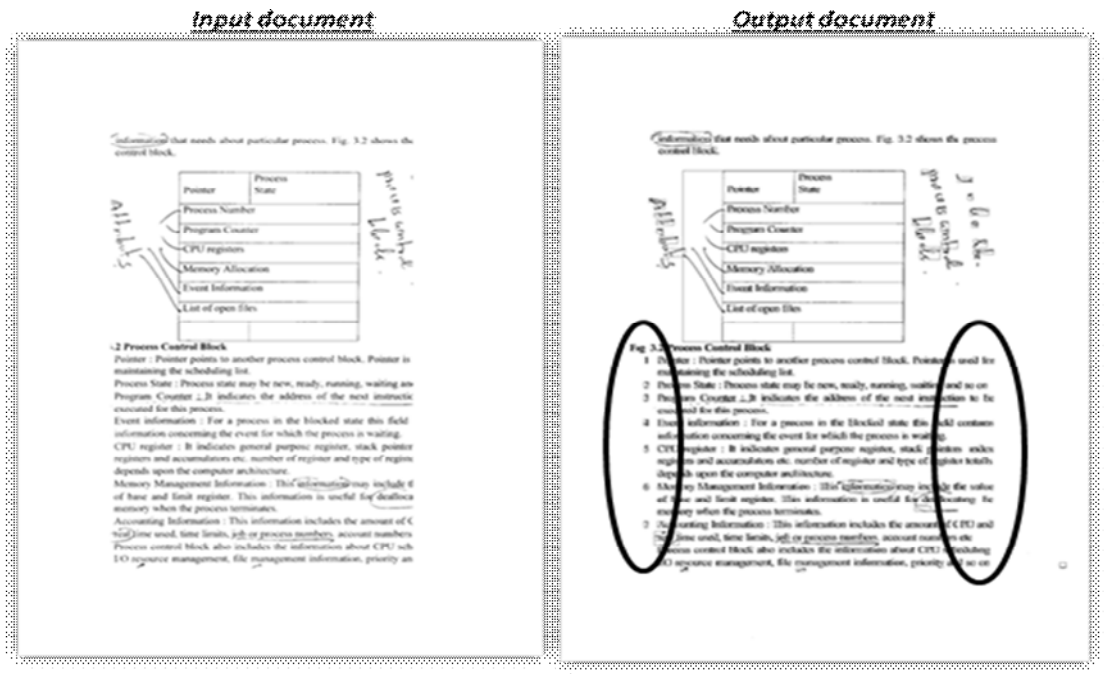

Fig 3.8 : Retrieving the printed text from the vertical marginal area to the body side

### *3.2.4 Retrieving page number*

Page number is a connected component which shares same characteristic as a character in the document body. In a standard paper, page number can be on the top/ bottom left/middle/right side of the document. To carry back the page number, we look for a connected component on the marginal side which satisfy two conditions: (i) It should be twice of character size according to our experimentation. (ii) It should respect the location of page number in standard document, the table below shows the strategy to find page number according to its probable location in a standard document.

Table-I: Location of page number in a standard document

| Probable location of page number | | Strategy to locate the page number and confirm its existence |
|---|---|---|
| In the y-axis (width of document) | Left | $0 < PageNumberZone < left\ margin$ |
| | Middle | $\frac{(leftmargin + rightmargin)}{2} - character_size < PageNumberZone$ and $PageNumberZone < \frac{(leftmargin + rightmargin)}{2} + character_size$ |
| | Right | $right\ margin < PageNumberZone < y$ |
| In the x-axis (length of document) | Top | $0 < PageNumberZone < Top\ margin$ |
| | Bottom | $Bottom\ margin < PageNumberZone < x$ |

Once the missed parts of the text lines are retrieved, some unwanted portions of connected components are also accompanied with them (for example some connected components have same character size as the printed text and belonging to the same horizontal position of the text line). In order to remove these unwanted components, we check for the presence of text components around it. In case the component has no text around it, then it is eliminated. Fig 3.9 shows removal of unwanted components.

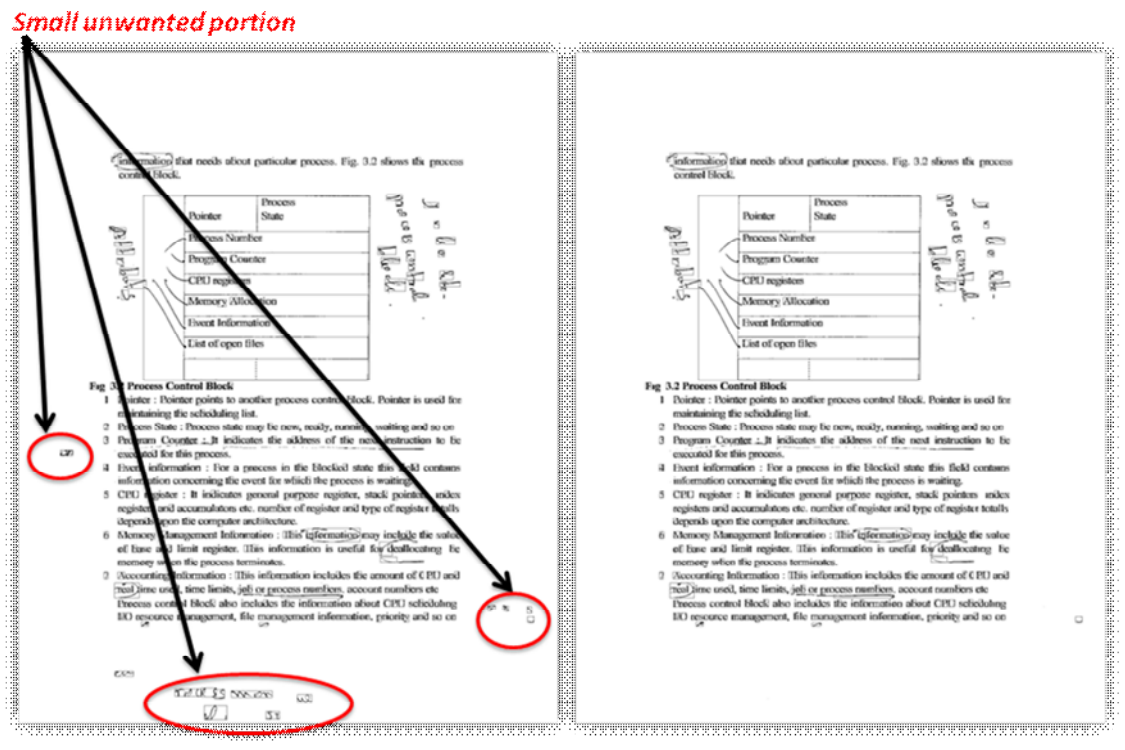

Fig 3.9 : Removing unwanted components

## 4. Experimental Result

To our best knowledge, we could not find any standard dataset with complex handwritten annotations, which suits our research requirements. Therefore, we have collected 50 documents from literary books and scientific articles, and given to heterogeneous group of students for marking random annotations of their choice. They generate the annotations in their own styles and using their own pens with different thickness. Dataset document images with annotations were obtained by using a scanner of 300 dpi.

We compute the accuracy of our proposed method in two ways.

### *4.1 Accuracy of removed handwritten annotations*

In order to compute the accuracy of removed marginal handwritten annotations we use the formula

$$Accuracy_{marginalAnnotations} = 1 - \frac{|B - A|}{A} \quad (3)$$

where ***A*** is the amount of expected marginal annotations to be removed, and ***B*** is the amount of removed marginal annotations. The experimental results show an average removal accuracy of 89.01%. Most of the documents in our dataset have accuracy above 90% as it is shown in Table-II.

Table-II: Experimental result of removing marginal annotations

| **Accuracy** | **Below 89%** | **Above 89.01%** |
|---|---|---|
| No. of Documents (%) | 38 | 62 |

However, it should be noted that the reduction in accuracy of 10.99% is observed not because of drawback of our proposed method, but due to our dataset which has been created for general annotations that can appear anywhere in the document and it was not restrictive to be entered in marginal space only. Therefore, our proposed algorithm removes any marginal annotations that enter the document body. The drop in accuracy of marginal annotations removal is observed because annotated document body is taken as ground truth during experimentation. Fig 4.1 demonstrates the case discussed above.

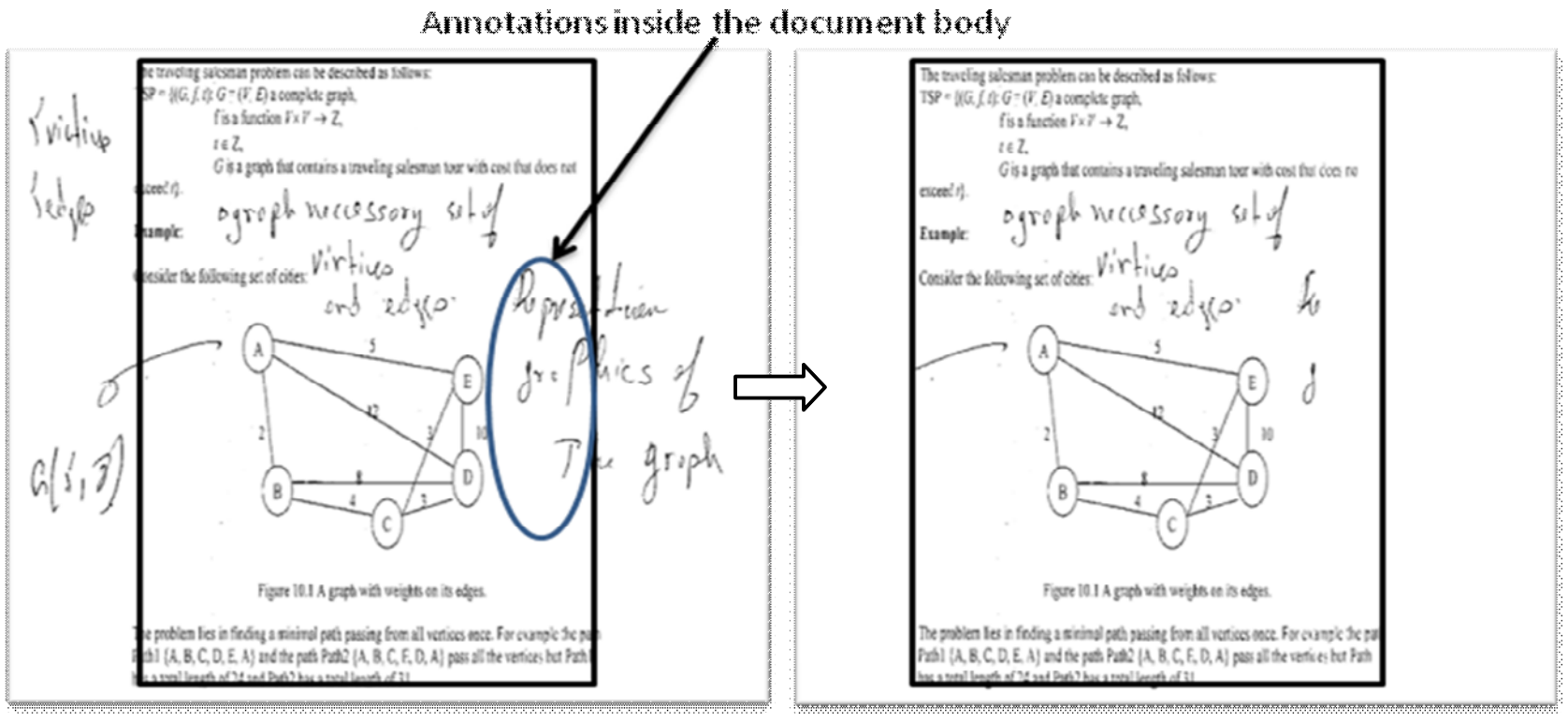

Fig 4.1 : Case of removal of annotations from inside the document body

### *4.2 Accuracy of recovering the original document*

To compute the accuracy of recovering the original document we use the formula

$$Accuracy_{originalDocument} = 1 - \frac{|\mathcal{B} - \mathcal{A}|}{\mathcal{A}} \quad (4)$$

where A is the expected cleaned document without marginal annotations, and B is recovered document after the removal of marginal annotations. Through our experiments, the average accuracy of getting the original document is 97.74% in which most of the documents of our dataset give an accuracy above the average as it is shown in Table-III.

Table-III: Experimental result of recovering original printed documents

| **Accuracy** | **Below 97.73%** | **Above 97.74%** |
|---|---|---|
| No. of Documents (%) | 24 | 76 |

We also compute the correlation coefficient between the original document and the processed document by using the formula

$$r = \frac{1}{n-1}\sum_{i=1}^{n}\left(\frac{x_i-\bar{x}}{s_x}\right)\left(\frac{y_i-\bar{y}}{s_y}\right) \quad (5)$$

where the number of pixels is given by $n$, the pixels of the original document are represented by $x_i$ and the pixels of the processed document are represented by $y_i$. The mean of the $x_i$, $y_i$ is denoted by $\bar{x}$ and $\bar{y}$. The standard deviation of the $x_i$, $y_i$ is denoted by $s_x$ ,$s_y$ respectively. However, the overall correlation between the original document and the processed document for 50 images of our database is showed on the Fig 4.2, with an average correlation of 0.9834.

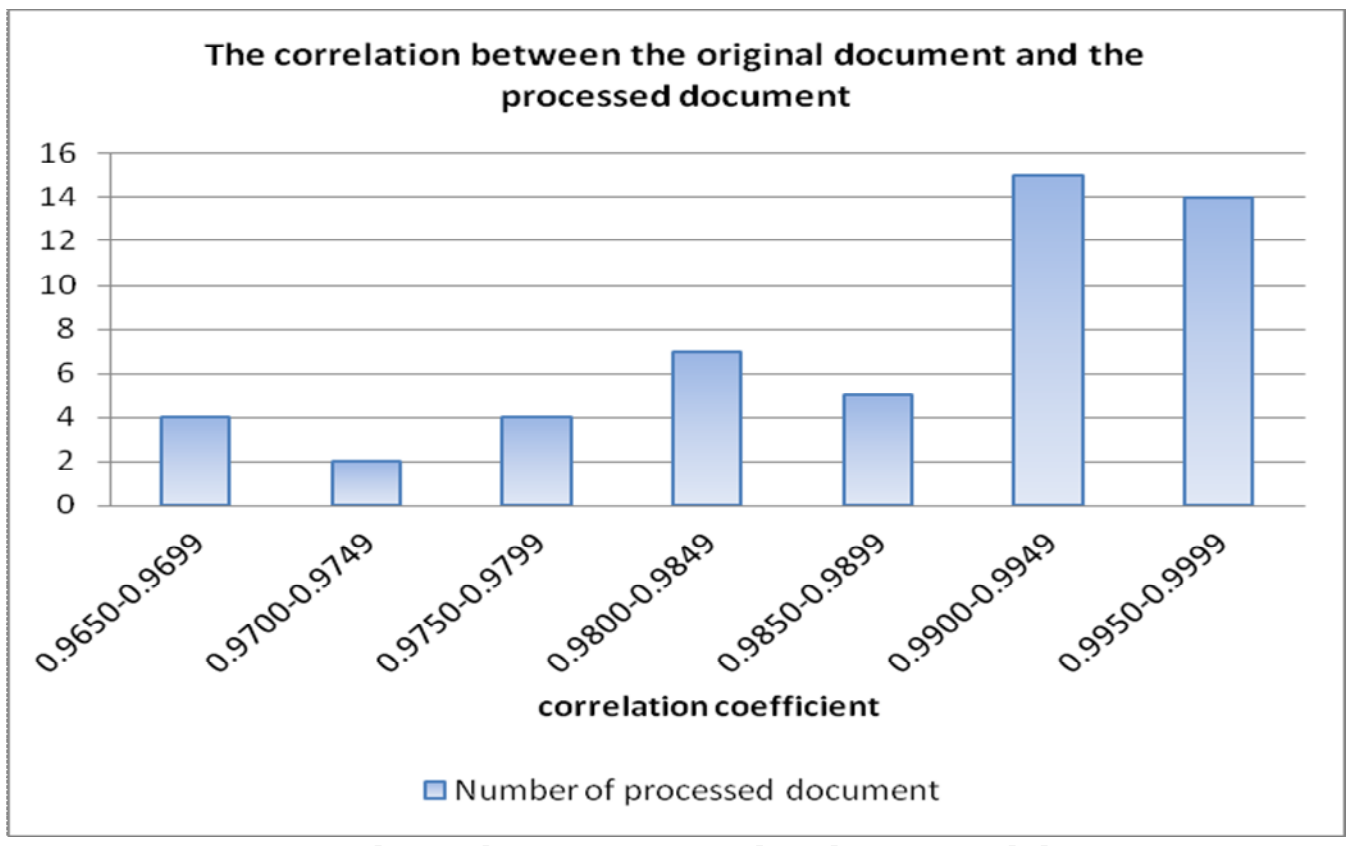

Fig 4.2: Correlation between original and processed documents

Therefore, it is clear that most of the documents which represent 82% from the dataset have a correlation above 0.9834.

Fig 4.3 demonstrates the relationship of correlation between the original document and the processed document with the accuracy of getting the original document from the annotated document. From the graph plot, we observe that the accuracy of getting the original printed text document is directly proportional to the correlation of the document.

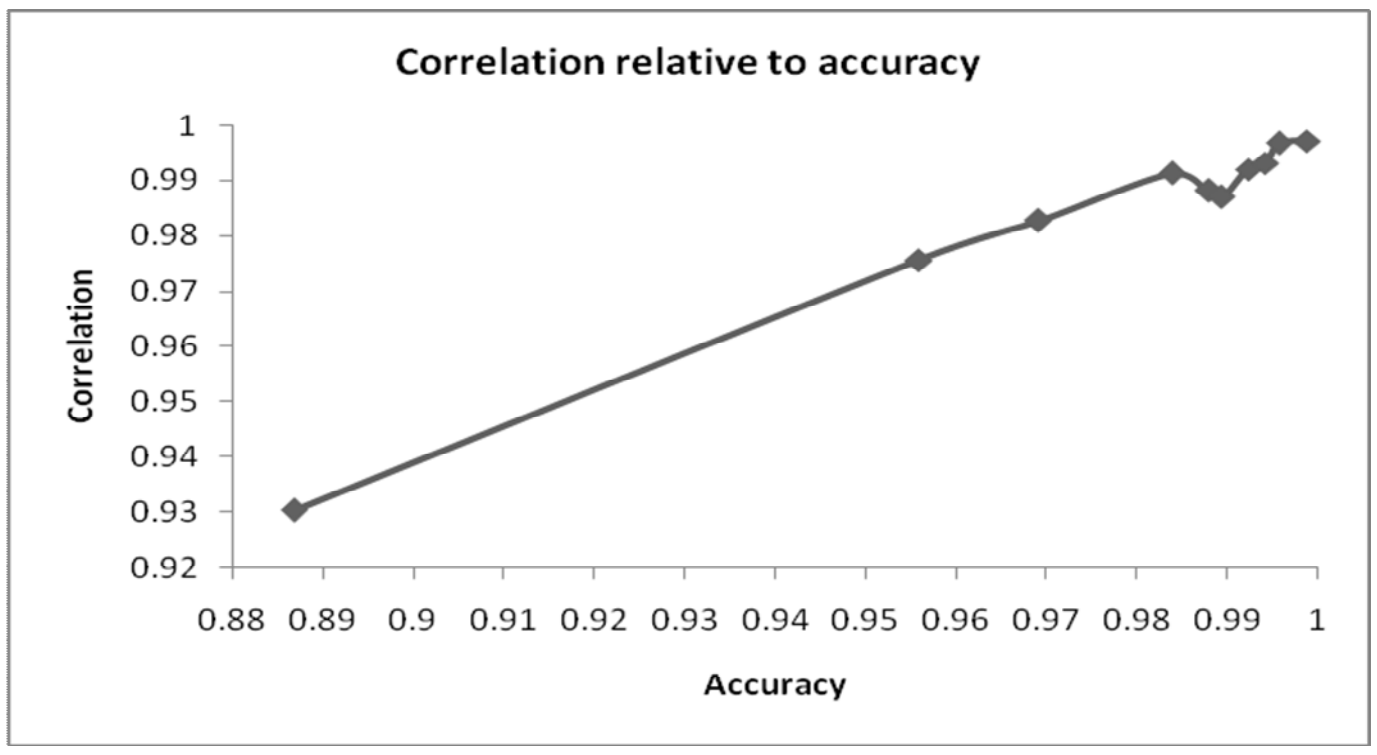

Fig 4.3: Correlation with respect to accuracy

## 5. CONCLUSION

In this paper, we proposed a novel method to remove marginal handwritten annotations from annotated printed text document. The experimental results show that our method produces consistent and adequate results under reasonable variation of type of annotations. We demonstrate the proposed idea on a dataset of 50 documents which gives an overall accuracy of 89.01% for removal of handwritten annotations, and 97.74% in case of recovering the original printed text document, with an average execution time of 6.8492 seconds implemented in MATLAB using 4GHz and 3GB RAM system.